# Inferring Taxi Status Using GPS Trajectories

Yin Zhu[+*], Yu Zheng[+], Liuhang Zhang[+], Darshan Santani[#], Xing Xie[+], Qiang Yang[*]

yuzheng@microsoft.com

[+] Microsoft Research Asia, China
[*] Hong Kong University of Science and Technology, Hong Kong
[#] ETH Zurich

**Abstract**

In this paper, we infer the statuses of a taxi, consisting of occupied, non-occupied and parked, in terms of its GPS trajectory. The status information can enable urban computing for improving a city's transportation systems and land use planning. In our solution, we first identify and extract a set of effective features incorporating the knowledge of a single trajectory, historical trajectories and geographic data like road network. Second, a parking status detection algorithm is devised to find parking places (from a given trajectory), dividing a trajectory into segments (i.e., sub-trajectories). Third, we propose a two-phase inference model to learn the status (occupied or non-occupied) of each point from a taxi segment. This model first uses the identified features to train a local probabilistic classifier and then carries out a Hidden Semi-Markov Model (HSMM) for globally considering long term travel patterns. We evaluated our method with a large-scale real-world trajectory dataset generated by 600 taxis, showing the advantages of our method over baselines.

# 1. Introduction

GPS-equipped taxis can be regarded as pervasive mobile sensors constantly probing a city's rhythm and pulse, such as traffic flows on road networks and city-wide mobility patterns of people. The trajectory data generated by these taxis imply rich and supportive knowledge that can enable urban computing for improving a city's transportation systems and/or devising an efficient land use policy [Zheng and Zhou 2011b].

To achieve the above scenarios, we first need to infer the statuses of a taxi, i.e. parked, occupied, or non-occupied by passengers, at a given time from its GPS trajectory. Here, the status of occupied ($O$) is defined as a taxi sending passengers to a destination, while the non-occupied ($N$) is referred to as a taxi cruising without passengers. Note that the definition of parked ($P$) is a location where a taxi has stayed and/or queued for a while with the intention to get a passenger on-board, or the situation when the taxi driver is taking a break (e.g., for lunch, coffee or going to a restroom). Using Figure 1 as an example, we present the differences between with and without the status of a taxi. As shown in Fig. 1 A), we have no idea what happened during the trip (from S to D) given the raw trajectory of a taxi. However, after the inference, we find the taxi first sent passengers to the location X, and then drove to a parking place in a status of non-occupied. After waiting for a while, the taxi took another passenger(s) to the location Y, and maintained its non-occupied state until reaching the place Z, where it picked up passengers again.

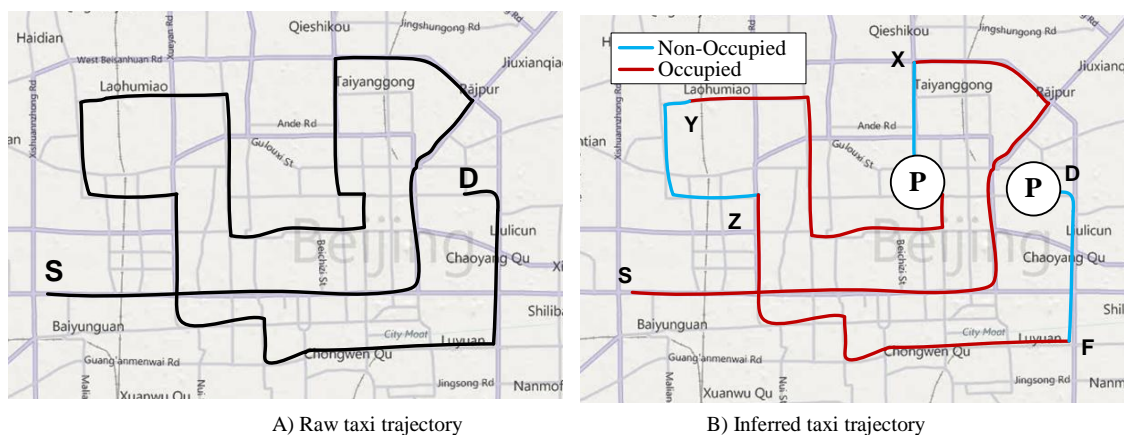

A) Raw taxi trajectory    B) Inferred taxi trajectory
**Figure** 1. The difference between with and without taxi statuses

The state information of a taxi contributes to the following three aspects at least. First, we can identify the real mobility of people and model the cite-wide mobility patterns more accurately [Zheng et al. 2011a]. Second, we are able to estimate the real-time traffic flow on a road surface more precisely by only using the speed information of taxis with passengers. Looking for passengers along streets, non-occupied taxis usually cruise slower than being occupied even if the traffic condition allows them to travel quickly [Yuan et al. 2010b]. Third, taxi operators can better manage their fleet by saving energy con-sumption and resources. For example, the operator can send a passenger's pick-up request to the non-occupied taxis, and identify high-performance drivers from their trajectories [Yuan et al. 2010a].

Though a few taxis built recently link a GPS sensor to a meter (so as to identify the taxi's status), these two devices remain largely unconnected in most taxis. As a result, we only have a small set of trajectories with labels while a lot of taxi trajectories remain un-labeled and still being generated continuously. According to our study, the scale of the labeled data is not sufficient to enable urban computing directly. Moreover, to be cost-effective, we do not necessarily connect these two devices for each taxi in the future. Having a small portion of taxis with these two devices connected, we can infer the states of other taxis.

In this paper, we aim to infer the status of a taxi ($O$, $N$, $P$) according to its GPS trajectory. The challenges of this work include: 1) *Low-sampling-rate and data sparse*: In practice, to save energy consumption a taxi usually updates their location every minute or more. As a result, it is common to have two consecutive GPS points generated on two different road segments (which could be far away from each other). Meanwhile, in the real-world, we can only have a limited and sparse trajectory dataset with state labels. 2) *Uncertainty*: A taxi trajectory could contain multiple trips sending different passengers to different places. The change of status depends on multiple factors including passengers, taxi drivers, time and locations.

Aiming to address these challenges, our work contributes the following aspects:

- Mapping a GPS point to a road segment, we identify and extract a set of effective features incorporating the knowledge of a single trajectory, historical trajectories of multiple taxis and geographic data, like road network and points of interest (POIs). The novel features and extraction method are useful in handling the first challenges by reducing the information space.

- We propose a two-phase inference model that first uses the identified features to train a local probabilistic classifier and then globally considers travel patterns via a Hidden Semi-Markov Model. This model well handles the second challenges.

- We evaluate our methodology using a large-scale real taxi dataset generated by 600 taxis in a period of 3 months. The results show the advantage of our approach over competing methods, such as HMM and CRF.

## 2. Related Work

To the best of our knowledge, no work has been done on directly inferring the taxi status. Our work is related to two areas:

Activity recognition from location trajectories. [Liao et al. 2007] predict an individual's transportation routine and recognize user-specific activities at each location, given the person's GPS data. [Zheng et al. 2008a; 2008b; 2010] infer the transportation modes from GPS trajectories. [Yin et al. 2008] do the indoor activity recognition using location trajectories inferred from WiFi Signals. [Zhu et al. 2011] provide an up-to-date survey on trajectory-based activity recognition.

Taxi trajectory mining. Zheng and Zhou [2011b] give a good tutorial on computing with spatial trajectories. For example, to optimize taxi drivers' income, [Ge et al. 2010] and [Liu et al. 2010] propose route recommendation services for a taxi driver by analyzing fleet trajectories. Besides the recommendation for a taxi driver, [Yuan et al. 2011] present a recommendation system that also provides people with some road segments (around their current position) where they can find a vacant taxi easily. [Yuan et al. 2010 and 2011] propose a smart driving directions recommendation service using the intelligence learnt from taxi trajectories. [Liu et al. 2011] and [Pang et al. 2011] aim to detect the anomalies in a city using GPS trajectories of taxis. [Zheng et al. 2011] try to identify the problematic urban planning using the traffic represented by taxi flows. All these research projects need to know the status of a taxi. [Phithakkitnukoon et al. 2010] is close to our work, but it aims to predict the number of vacant taxis near a given location rather than directly modeling taxi trajectories.

Our decision tree + HSMM model is similar to the hybrid classification framework in (Lester et al. 2005). However we focus more on the local decision + global smoothing with long-term sequential patterns rather than hybriding discriminative and generative classifiers.

## 3. Methodology

We first detect parking places from a taxi trajectory using our parking place detection algorithm, and then split a trajectory into some segments, which are sub-trajectories between two consecutive parking statuses. Later, we use the proposed two-phase inference model to infer the status, O or N, of each point from a taxi segment.

Before going to the details of our method, we first introduce the notations and formulate our problem. Our dataset contains a set of taxis $U = \{u_i\}_{i=1}^{N}$; each taxi $u_i$ has a set of trajectories $\{T_j^{(i)}\}_{j=1}^{M_i}$. A trajectory $T_j^{(i)}$ is a labeled GPS sequence $\{x_k, y_k\}_{k=1}^{|T_j^{(i)}|}$, where $x_k$ represents the k-th GPS point in the trajectory and $y_k$ is its status label. Each GPS point consists of three values: latitude, longitude and timestamp. Given a set of taxis with labeled trajectories as the training data, our goal is to build a model, which infers the status of each GPS point from tested taxis. Note the important setting that the training set and the testing set are from different taxis because the primary application of this work is to inference the status for cars that are not equipped with meter-linked GPS.

## 3.1 Parking Place Detection

We treat parking place detection as a standalone procedure as parking status is different from the behaviors of a running taxi, and can have additional applications, e.g. hot place detection in a city. In this step, we first detect parking place candidates using the density-based clustering algorithm shown in Figure 2 A-D). Then, we remove some false candidates caused by traffic jams or traffic lights from the candidate set, using a supervised model which incorporates the features illustrated in Figure 2 E-F).

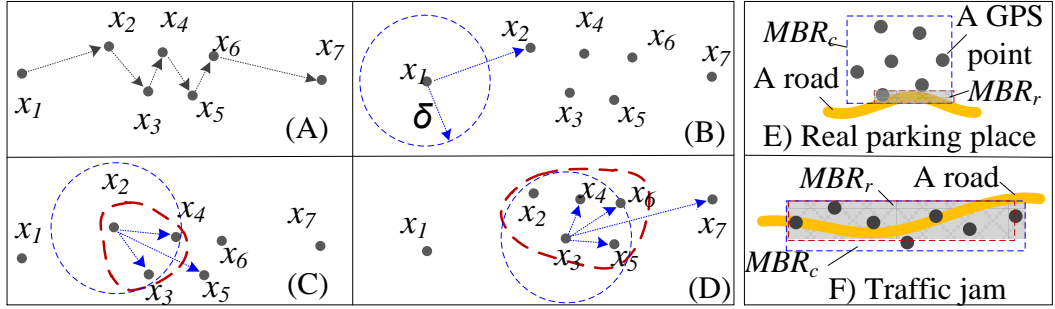

**Figure** 2. Illustration of the parking candidate selection algorithm.

An important observation of parking status is that the parking status spans over a set of very near GPS points. Thus we first use a candidate selection method to select these sets of points with locality: for a trajectory $T = \{x_k\}_{k=1}^n$, if $dist(x_i, x_{i+j}) < \delta$ for $1 \leq j \leq m$ and $timeDiff(x_{i+m}, x_i) > \tau$ (we call $x_i$ pivot point, $m$ is the maximum number of satisfied conditions and $\delta$ and $\tau$ are threshold parameters.), then we find a candidate parking set $C \{x_i, ..., x_{i+m}\}$. With this set in hand, we then try to grow it using $x_{i+1}$ as the next pivot point (See Figure 2(C&D)) until we cannot further expand the set under the two constraints introduced previously. This growing candidate algorithm is reminiscent of Breadth First Search algorithms and density-based clustering (Sander et al. 1998). We repeat this process until all the points in the trajectory have been scanned, and get the parking candidate sets $\{C_i\}_{i=1}^K$ for the trajectory $T$.

Essentially, the candidate detection algorithm finds out the locations where the GPS points of a taxi are densely clustered, with spatial, temporal and speed constraints. However, a parking place candidate could sometimes be generated by taxis stuck in a traffic jams, or waiting for signals at a traffic light, instead of a real parking.

To reduce such false selections, we design a supervised model for picking out the true parking places from the candidate sets, using the following features: 1) *Minimum Bounding Ratio (MBR).* As shown in Figure 2(E&D), MBR is the area ratio between the bounding box of the road segment (MBRr) and the bounding box of the GPS points (MBRc) in the candidate set. 2) *AverageDistance*. The average distance between points in the candidate set and their nearest road segments. 3) *CenterDistance*. The distance between center point of MBRc of the candidate set and the road segments. 4) *Duration*. The parking dura-tion m. 5) *History*. The number of parking candidates within 50 meters in the past 7 days. 6) *PoiVector*. This is a set of numbers with each representing the number of specific POI within 50 meters.

## 3.2 Occupied and Non-occupied Inference

How could one know whether a running taxi is occupied by looking at its trajectory? Humans could do this by experience in the form of heuristics. For example, the geometric trajectory pattern when a taxi driver is trying to find a passenger is usually different from that of an occupied taxi. As another heuristic, a taxi seldom drops off or takes on passengers on a high-way. The mathematical modeling in this section is to formulate these heuristics in a principled way.

### 3.2.1 Probabilistic Decision Tree for Local Inference

For each GPS point in a trajectory, we extract features that are relevant to the status detection. These features encode the in-domain heuristics that we think are potential indicators of the taxi status. We then follow standard supervised learning setting to use these features and the corresponding status labels to train a model. Among various supervised learning methods that can be applied here, we choose decision tree with probabilistic outputs as our classifier for both its effectiveness and efficiency. Decision trees also have the virtue for its well-interpretable tree models, which tell back what are the most effective heuristics the machine "thinks".

Aside from the trajectory information, we also use two external knowledge sources: the road network and POI (Point of Interest) information. We group the extracted features of a GPS point into three categories (the three categories are also shown in Figure 3(A) :

1. *Features based on the trajectory alone*. We set a window size at one GPS point, and calculate the speed and the ratio between direct line distance and the actual trajectory distance. The window size varies from 1 to 4. The time stamp is also extracted as a feature.
2. *Enrich raw GPS trajectory with road map and POI (Point of Interest) information*. We do map matching for GPS points in every trajectory by a scalable method [Yuan et al. 2010a]. We also find the nearby POIs around a GPS point within a square bounding box of size 50 meters centered at that GPS point. Thus a GPS point in the taxi trajectory has its associated Road ID, and counts for nearby POIs. The POI categories we select are "4&5 stars hotels", "parking", "shopping" and "entertainment". We also extract the road ID from the previous GPS point in the trajectory. In this way, road transition could be encoded in features.
3. *Statistics from historical trajectories*. For each road, we calculate the portion that taxis are occupied on it conditioned on timestamps. We also calculate the status transition probability between different road segments.

The road network puts constraints on the layout of the GPS points – the mapping from a GPS point to its road id could be viewed as information compression. Originally the possible transition between GPS points lies in an infinite space, now it reduces to a discrete and finite road network. Thus these features can effectively solve the low-sampling-rate and data sparsity problem.

In total, we have 26 different features extracted from each GPS point. In the experiment part, we will show which of them are the most effective features.

We use the calibrated C4.5 decision tree in [Bianca Zadrozny & Elkan 2001] as our probabilistic decision tree implementation. As described in the supervised learning benchmark work by [Caruana & Niculescu-Mizil 2006], calibrating the probability output of a standard classifier using a single variable sigmoid function can lead to better probabilistic output [Platt 1999]. Suppose the probability of GPS point $x_i$ being occupied is $f_i$, then the calibrated probability is

$$p_i = \frac{1}{1 + \exp(Af_i + B)} \ ,$$

where parameters A and B are obtained by fitting the single variable logistic regression model.

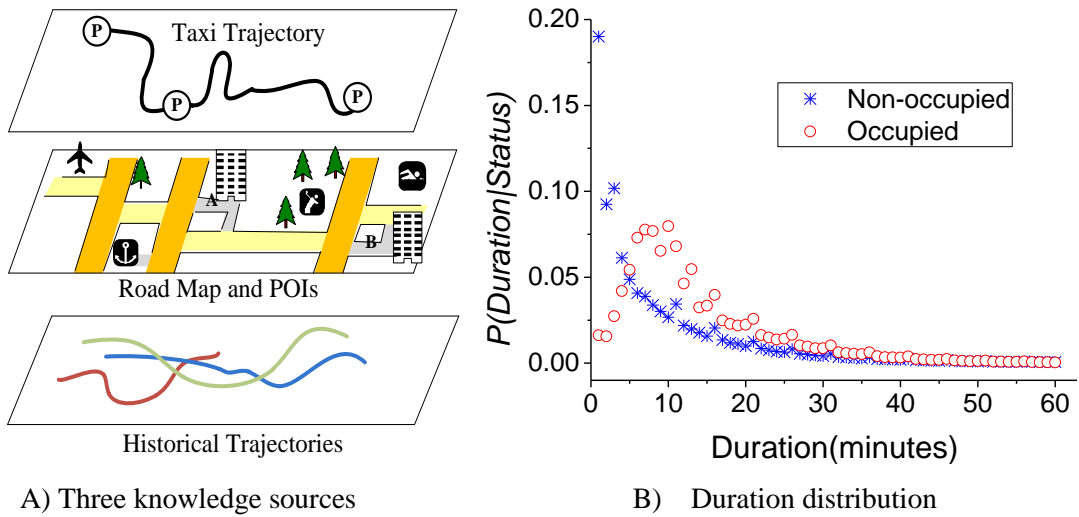

A) Three knowledge sources   B) Duration distribution

Figure 3. Knowledge & patterns for inferring taxi status.

### 3.2.2 HSMMs for Global Smoothing with Travel Patterns

A non-parking taxi trajectory is constituted by consecutively and interleavingly connected occupied and non-occupied sub-trajectories. One important observation in these trajectories is that the occupied duration and non-occupied duration follow consistent patterns. We collect the P(duration|status) distribution from 600 taxis and draw the distribution in Figure 3(B). As shown in the figure, there is 20% chance for a taxi driver to find a new passenger within one minute after dropping off its previous passenger; and the duration distribution for occupied segments peaks at around 10 minutes.

Hidden Semi-Markov Models (HSMMs) [Yu 2010] extend the standard HMMs [Rabiner 1989] by allowing each hidden state to associate with a variable duration or sojourn time, which is also called the Semi-Markov property. Besides the hidden states and their (emitted) observations defined in HMMs, HSMMs have an explicit duration d for the hidden state. In the discrete case, HSMMs have variable observations in one state while HMMs only have one observation in each state. The duration for each possible hidden state follows a discrete distribution, which exactly models the travel pattern for occupied and non-occupied states in our taxi domain.

pplied in our taxi domain, we define the hidden state set $Q = \{O, N\}$. The observation for each GPS point is the probability of its being labeled as occupied as calculated by the decision tree. We discretize the probability into B bins and use discrete set $O = \{O_1, O_2, \ldots, O_B\}$ to represent the observation set. The duration is a random variable taking values from integer set $D = \{1, 2, \ldots, D\}$.. Here $B$ and $D$ are model parameters that should be provided. When $D=1$, the model degenerates to a standard HMM.

A HSMM model $\lambda$ is defined as $\lambda \triangleq \{a_{(i,d')(j,d)}, b_{j,d}(v_{k_1:k_d}), \pi_{i,d}\}$,, where $a_{(i,d')(j,d)}$ is the state transition probability from state $i$ with duration $d$ to state $j$ with duration $d'$, $b_{j,d}(v_{k_1:k_d})$ is the emission probability from state j with duration d to its observation value $(v_{k_1:k_d})$, and $\pi_{i,d}$ is initial probability for the first observation in a sequence.

Similar to HMMs, the inference and estimation algorithms for HSMMs also involve the calculation of forward and backward variables except that HSMMs have the duration parameter d while HMMs only have a state parameter. The forward and backward variables are defined as

$$\alpha_t(j,d) \triangleq P[Q_{[t-d+1:t]} = j, o_{1:t} \mid \lambda]$$
$$= \sum_{i \in S \setminus \{j\}} \sum_{d' \in D} \alpha_{t-d}(i,d') \cdot a_{(i,d')(j,d)} \cdot b_{j,d}(o_{t-d+1,t}),$$

and

$$\beta_t(j,d) \triangleq P[o_{t+1:T} \mid Q_{[t-d+1:t]} = j, \lambda].$$
$$= \sum_{i \in S \setminus \{j\}} \sum_{d' \in D} a_{(j,d)(i,d')} \cdot b_{i,d'}(o_{t+1,t+d'}) \cdot \beta_{t+d'}(i,d').$$

The inference algorithm for HSMMs is based on the forward and backward variabels. The maximum a posteriori (MAP) estimate of state $s_t$ given a specific observation sequence $o_{1:t}$ can be obtained by maximing

$$\hat{s}_t = \arg\max_{i \in S} \{\gamma_t(i)\}$$

and $\gamma_t(i)$ is calculated based on forward and backward variables:

$$\gamma_t(i) = \sum_{\tau \geq t} \sum_{d=\tau-t+1} \alpha_t(j,d)\beta_t(j,d)$$

To speed up the dynamic programming decoding procedure, we only allow two kinds of state transitions: $(i,d) \to (j,1)$ for $i \neq j$ and $(i,d) \to (i,d+1)$ for self-transitions. The computational complexity for inference is $(|\mathbf{Q}|^2|\mathbf{D}|T)$, which is very fast for online inference.

Estimating the model parameter λ involves an initialization from the training data and an EM procedure for updating λ. Due to space limit, we refer interested reader to [Yu 2010] for details.

### 3.3 Discussing Alternative Models

The taxi status inferring problem could also be treated as a standard sequential classification task, for which standard models such as Hidden Markov Models (HMMs) and Conditional Random Fields (CRFs) could be applied. In this subsection, we discuss how to use HMMs and CRFs and some of their variants to model our taxi status inferring problem.

Using HMM to our problem is straightforward; one problem is that the observation in each GPS point is actually a feature vector. We could either model it as a multivariable Gaussian or discretize it into a discrete value.

We prefer the later as the observation does not follow Gaussian distribution and discrete observation HMM usually performs better. The disadvantage of HMM is that the transitions between observations (e.g. road transitions) could not be modeled explicitly [Klinger & Tomanek 2007]. We overcome this shortcoming of HMMs by representing the transitions directly as input features for local decision trees, which could deal with large number of features easily while HMMs suffer from the large observation/feature space.

Conditional Random Fields (CRFs) provide more flexible modeling over HMMs in our problem as it allows direct relationship modeling on the observations. For example, the road transition in a taxi trajectory could be explicitly modeled by CRFs. However, how to represent the features remains a challenging problem. In linear-chain CRFs, the probability of the labeling y for an observation sequence x is represented as

$$p(\mathbf{y}|\mathbf{x}) = \frac{1}{Z(x)} \prod_{t=1}^{n} \Phi(t, y_t, y_{t-1}, \mathbf{x}),$$

where Φ is a vector of feature functions:

$$\Phi(\cdot) = [f_1(\cdot), f_2(\cdot), \ldots, f_K(\cdot)]^T$$

We follow the practice in the activity recognition work by (Mahdaviani & Choudhury 2007) in which binary tree stumps in the forms of $f_i(t, y_i, y_{i-1}, x) = y_i = s \,\&\&\, x_i > h$ are used as emission feature functions. However, the thresholds h for tree stumps could not be learned by CRFs.

CRFs also have a semi-Markov variant [Sarawagi and Cohen 2005]; however it could only be used to model short duration state, e.g. name entity recognition, where the name entity usually covers less than 5 consecutive words. The reason of this is that in semi-Markov CRFs, features are extracted from the whole segment with the same label. For long segments such as a 30-minute occupied taxi trajectory, it is very hard to design descriptive feature functions for them.

Thus the discussion leaves us using standard HMM and CRF as our baselines.

## 4. Experiments

### 4.1 Dataset

Our collaborator is a newly founded fleet operator in Beijing that has 8000 taxis equipped with meter-linked GPS devices [Yuan et al. 2010b; 2011]. Thus the GPS points in the trajectories are marked with binary status labels – occupied with passengers or not. We select 600 taxis as our experiment data from the 8000. This is only a small number of the total taxis in Beijing (Beijing has about 70,000 taxis but a lot of the taxis in the city are not equipped with meter-linked GPS.) The selected taxis have relative stable sampled trajectories. We cannot guarantee that the GPS sampling frequency is exactly one point per minute. But statistics shows that 83.4% of the GPS points from the 600 taxis are within 50 to 70 seconds to their previous points. In total, there are 161,349 trajectories with 25,410,290 GPS points. The time span of the dataset is 1.5 months, thus on average a taxi has about 7 trajectories and about 940 GPS points per day.

For the non-occupied points, we further divide them into two categories: parked and non-parked by manual labeling. Two locals are hired to label the trajectories from 20 taxis using a data visualization tool we developed in-house. We first use the parking candidate selection algorithm to mark the potential parking candidate sets out with very conservative parameters $\delta$=50 meters and $\tau$=3 minutes. This selects 510 candidate sets from the 20 taxis. Then each labeler label the candidate sets with one of the following three labels: I) parked with confidence, II) probably parked and III) non-parked with confidence. We select candidate sets with label I from either of the labeler s or label II from both of the labelers as parked. Among the 510 candidates set, 253 of them are labeled as parked.

### 4.2 Results

We first evaluate our method for parking detection. Because the parking detection is more close to an extraction task – finding the parking candidate sets out from others, we use F1-measure as the evaluation criterion. We also report the predication accuracy for reference. We perform 10-fold cross validation using C4.5 decision tree as the supervised learning classifier. The four features with the highest mutual information (MI) against the true labels are *MBR, Duration, History* and *#Hotels*. We delete each of them and re-evaluate cross validation process and show the comparing result in Table 1.

Table 1. Evaluation result for parking detection

| Features | All | No MBR | No Duration | No History | No #Hotels |
|---|---|---|---|---|---|
| F1-meas. | 0.858 | 0.848 | 0.702 | 0.837 | 0.853 |
| Accuracy | 0.864 | 0.854 | 0.741 | 0.845 | 0.860 |

For the next experiment, we compare our proposed occupied and non-occupied detection method with three baselines: decision tree (DT), HMM and CRF. First we use the parking detection model to filter all the parking points it detects in our dataset so that long trajectories are divided into short ones with parking segments in between; and we assume that all the trajectories left are from running taxis. We hold out 100 taxis (21,394 trajectories and 3,427,611 GPS points in total, 57.1% of the GPS points are marked as occupied) as our evaluation set. We vary the training data size from 100 to 500 from the rest 500 taxis. The experiment result is shown in Figure 4(A). The parameters for HSMM are set as D=20 and B=100. As we can see, our DT+HSMM method consistently outperforms other baselines. To our surprise, DT has better performance than HMM and CRF. Although DT uses the same feature set as HMM and CRF, the latter two cannot utilize the full power of the extracted features. In HMM, there is a discretization in the features. In CRF, the thresholds for the tree-stump style feature functions cannot be learned while decision tree has very good split between feature values via information theory based training. We also study the longest duration D in HSMM. Figure 4(B) shows the accuracies when varying the duration parameter D from 1 to 40 (The training size is 500 taxis and other parameters are also fixed). The accuracy converges when the duration parameter D≥20. Note that when D=1, the model is actually a DT+HMM, which performs worse than DT.

We have also printed the decision tree out and found that the following features are at the top of the tree: *roadLevel*, *roadVote* and *distance3*. Interestingly, each feature category has one representative in the top three. Feature *roadLevel* comes from the road network knowledge source. It has four values indicating the road types. Feature *roadVote* comes from historical statistics category, and it measures the portion of occupied GPS points in the road in the whole training data. Feature *distance3* comes from the trajectory information category, and it is the distance between the current point $x_i$ with the previous point $x_{i-3}$.

We also do a small experiment to put together the three statues. Figure 4(C) shows the confusion matrix for the 10-fold CV result of the 20 taxis with parking labels.

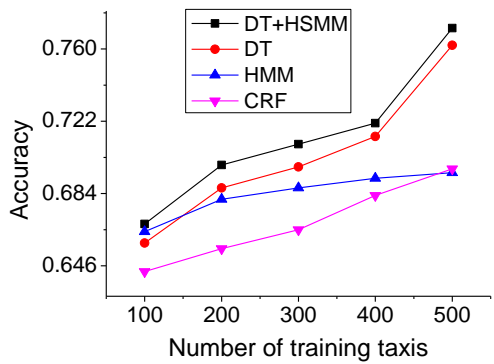 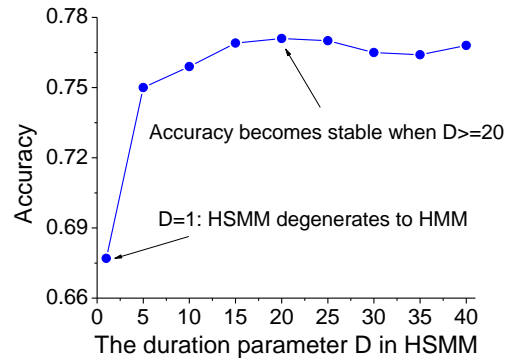

A) Algorithm comparison.   B) Varying in parameter D.

|   | *N* | *O* | *P* |   |   |
|---|---|---|---|---|---|
| *N* | 12648 | 4387 | 213 | 0.733 | Recall |
| *O* | 5372 | 12212 | 662 | 0.669 |  |
| *P* | 121 | 261 | 3672 | 0.905 |  |
|   | 0.697 | 0.724 | 0.807 | Accuracy: 0.721 |  |
|   | Precision |  |  |  |  |

C)  Confusion matrix for 20 taxis.
**Figure 4.** Experimental results**.**

## 5. Conclusion

In this paper, we have proposed a novel decision tree + HSMM approach for taxi status inference which outperforms other methods including HMMs and CRFs. With efficient training and inference phases, our model well encodes the local features and global travel patterns. The proposed method handles the challenges of low-sampling-rate, data sparsity and the uncertainty among status transitions. The inferred statuses have a broad range of applications in urban computing, such as transportation systems and land use planning.

In the future, we will deploy our model into real taxi systems. We will also consider non-consistent sampling trajectories, and more broadly we will mine detailed activities from taxi trajectories such as "trying to find a passenger" or "going to lunch and don't carry passengers".